\documentclass[sigconf]{acmart}

\acmConference[C+J25]{Computation + Journalism Symposium}{2025}{Miami, Florida, USA}

\setcopyright{none}
\usepackage{enumitem}
\setlist[itemize]{topsep=0pt, partopsep=0pt, itemsep=0pt, parsep=0pt}

\makeatletter
\renewcommand\footnotetextcopyrightpermission[1]{}
\makeatother
\setcopyright{none}

\acmConference[C+J '25]{Computation + Journalism Symposium}{Dec 11--12,
  2025}{Miami, FL, USA}

\begin{document}

\title{Impacts of Racial Bias in Historical Training Data for News AI}

\author{Rahul Bhargava}
\email{r.bhargava@northeasten.edu}
\orcid{0000-0003-3904-4302}
\affiliation{%
  \institution{Northeastern University}
  \city{Boston}
  \country{USA}
}

\author{Malene Hornstrup Jespersen}
\email{malenehj@sodas.ku.dk}
\affiliation{%
  \institution{University of Copenhagen}
  \city{Copenhagen}
  \country{Denmark}
}

\author{Emily Boardman Ndulue}
\email{ebndulue@mediaecosystems.org}
\orcid{0000-0001-9859-1627}
\affiliation{%
  \institution{Media Ecosystems Analysis Group}
  \city{Boston}
  \country{USA}
}

\author{Vivica Dsouza}
\email{dsouza.viv@northeastern.edu}
\affiliation{%
 \institution{Northeastern University}
  \city{Boston}
 \country{USA}
}



\begin{abstract}

AI technologies have rapidly moved into business and research applications that involve large text corpora, including computational journalism research and newsroom settings. These models, trained on extant data from various sources, can be conceptualized as historical artifacts that encode decades-old attitudes and stereotypes. This paper investigates one such example trained on the broadly-used New York Times Annotated Corpus to create a multi-label classifier. Our use in research settings surfaced the concerning \textit{blacks} thematic topic label. Through quantitative and qualitative means we investigate this label’s use in the training corpus, what concepts it might be encoding in the trained classifier, and how those concepts impact our model use. Via the application of explainable AI methods, we find that the \textit{blacks} label operates partially as a general “racism detector” across some minoritized groups. However, it performs poorly against expectations on modern examples such as COVID-19 era anti-Asian hate stories, and reporting on the Black Lives Matter movement. This case study of interrogating embedded biases in a model reveals how similar applications in newsroom settings can lead to unexpected outputs that could impact a wide variety of potential uses of any large language model–story discovery, audience targeting, summarization, etc. The fundamental tension this exposes for newsrooms is how to adopt AI-enabled workflow tools while reducing the risk of reproducing historical biases in news coverage.

\end{abstract}

\keywords{algorithmic auditing, AI in newsrooms, editorial algorithms, news classification, bias in journalism, computational journalism ethics}




\ccsdesc[500]{Computing methodologies~Machine learning}
\ccsdesc[500]{Social and professional topics~Computing / technology policy}
\ccsdesc[300]{Human-centered computing~Collaborative and social computing}

\keywords{algorithmic auditing, AI in newsrooms, editorial algorithms}

\maketitle

\section{Introduction}

Newsrooms are rapidly integrating a variety of AI tools into their reporting, writing, and publishing workflows. AI tools promise to improve tasks such as story recommendation, story customization, content summarization, and more. At the center of these tools is the representation of word usage probabilities as high-dimension vectors via a large language model (LLM). However, beneath this veneer of technological sophistication lies the conceptualization of these futuristic “things” as historical artifacts that encode attitudes, stereotypes, and language norms of their training data \cite{suchman_uncontroversial_2023}.

The field of algorithmic auditing offers techniques to understand this kind of historical bias that can be applied in news-related settings. While definitions of “bias” vary \cite{varona_discrimination_2022}, here we engage the particular definition that relates to human prejudices embedded in training data that are encoded into a model. In this paper we explore related concerns through a case study of a multi-label classifier trained on the New York Times Annotated Corpus \cite{sandhaus_new_2008}; the classifier model further used the Google News word2vec \cite{mikolov_distributed_2013} as the vectorizer. While situated in a journalism research context as opposed to in a newsroom, the technical task and software architecture are analogous to similar text classifiers used in reporting, editing, recommending, and other publishing areas. 

In usage, this classifier revealed a potentially problematic thematic topic label: \textit{blacks}. To a contemporary reader, this descriptive for African Americans or Black Americans sounds antiquated and othering. Using a variety of auditing methods, we found that the label encodes racial attitudes from previous decades that could systematically distort contemporary news analysis tools. Through this case study we demonstrate two critical insights that have broad implications for AI-based computational journalism: first, that \textbf{historical training data can fundamentally misrepresent the social categories it claims to classify}, and second, that \textbf{temporal gaps between training and application create systematic oversights that pose risks to newsroom AI systems}. Understanding and addressing these are essential for ensuring that journalism's technological evolution supports rather than undermines its democratic mission. This case study contributes to our growing understanding of this challenge, offering one example of how that can manifest in reproducing historical prejudices, and documents a set of methods one could undertake to assess these potential impacts.

\section{Related Work}

\subsection{AI in Research and Newsroom Contexts}

In research domains, scholars have widely adopted and discussed the role classifiers can play in understanding large corpora of news text \cite{barbera_automated_2021}. Various work has surveyed opportunities and challenges related to newsrooms integration \cite{de_grove_what_2020,taha_comprehensive_2024}.

In newsroom domains, regular reports in industry trade publications document applications in newsrooms small and large \cite{bhatia_how_2015,lichterman_ap_2016}. The domains include fact-checking, summarization, personalization, and beyond. While some work explores potential biases and dangers, the question of concrete impacts on news analysis and production is often unexplored in depth. Some contemporary organizational leaders believe that using AI classifiers on news can help eliminate bias \cite{folkenflik_l_2024,deck_law360_2025}.

\subsection{Historical Bias in Training LLMs}

Many ML applications operate as “black boxes” \cite{ojewale_towards_2024} where the implications of misinterpretations can lead to miscommunication or misinformation \cite{hofmann_ai_2024}. “Post-hoc” auditing is one approach that can help understand a model that has already been developed and deployed \cite{madsen_post-hoc_2023}. These tools attempt to discern the behavior of complex models via proxies that are more understandable to human researchers \cite{hooshyar_problems_2024}.

Over the last few years there has also been notable attention paid to the question of how LLMs become outdated. Computation scholars have found evidence of “temporal bias” in LLMs, and explored various techniques for adding in contemporary context \cite{wallat_temporal_2024,zhu_is_2025}. More relevant for this case study, Mozzherina documented and suggested remedies for temporal inconsistencies in the NYT Annotated Corpus, specifically. Their work reassigned labels based on clustering and iterative evaluation, yielding reduced redundancy in tags and increasing consistency \cite{mozzherina_approach_2013}.

\subsection{Media and Minoritization}
Our focus on the racially loaded label \textit{blacks} for this study requires an understanding of the historical norms about labeling and reporting on this minoritized group by U.S. media. While a thorough exposition of these norms is beyond the scope of this paper, negative portrayals of Black Americans in US media have historically exacerbated harmful stereotypes \cite{kumah-abiwu_media_2020,matsa_black_2023}, and language used in the media to refer to Black people has changed over time \cite{smith_changing_1992,niven_elite_2000,wihbey_racial_2015}. Recently, many news organizations have revised their style guidelines and reporting practices in response to evolving social norms and calls for more inclusive, accurate representation \cite{evans_if_2019, nguyen_recognizing_2020}. It is critical that new technologies used in newsrooms allow for journalists to stay current with preferred language and ways of referring to and representing historically minoritized groups, while still staying true to their core journalistic values.

\section{Methods}

Understanding the use and impact of the \textit{blacks} label in our thematic classifier required mixing qualitative and quantitative methods. We created four datasets of seen and unseen data to use as evaluation sets:

\begin{itemize}
  \item \textbf{Evaluation Set A}: A random subset of 5,000 articles from the training data labeled \textit{blacks} in the NYT corpus. We expect these to reflect concepts that the label encodes.
  \item \textbf{Evaluation Set B}: A random subset of 5,000 articles from the training data not labeled \textit{blacks} in the NYT corpus. We expect these to reflect concepts the label does not encode.
  \item \textbf{Evaluation Set C}: All articles from a curated collection of “Black US Media” sources in April 2023 (437 articles). This collection was sourced from the \hyperlink{https://www.journalism.cuny.edu/2020/11/mapping-black-media/}{Mapping Black Media project}, including US-based news sources primarily serving Black communities. This corpus represents a set of topics those editors believe are relevant to the Black community, and thus we expect them from a naive perspective to be strongly labeled \textit{blacks} by the model. 
  \item \textbf{Evaluation Set D}: All articles from a curated collection of “US National” sources in April 2023 (8,163 articles). This is a manually curated list of sources with broad national coverage, such as the Washington Post and USA Today. It effectively models a spectrum of data we naively expect to have a similar distribution of \textit{blacks} label scores when compared to the original training data.
\end{itemize}

\subsection{Quantitative Analysis: Explainable AI}

We sought to understand concepts the \textit{blacks} label encodes by using LIME \cite{ribeiro_why_2016}. LIME attempts to faithfully produce a post-hoc interpretable explanation of a prediction by making small changes to the input text fed into a model. With LIME, we can thereby attempt to answer questions such as "which terms had highest importance for this prediction?" We ran LIME analysis via 1,000 tweaked samples on twenty random articles scoring above 0.2 probability as \textit{blacks} from each of the four evaluation datasets. This probability threshold was previously determined to be a useful level for displaying labels within our research tool. We ignore other label outputs to focus on the binary prediction of the label \textit{blacks} or not.

\subsection{Qualitative Analysis: Content analysis}

Analyzing most frequent words does little to reveal the contexts those words were used within. We conducted two forms of qualitative analysis in parallel to gain an in-depth understanding of which articles the model predicted with highest probability to contain the label \textit{blacks}.

Our first approach involved reading the 25 articles from each evaluation set with the highest predicted scores for the \textit{blacks} label. For each article read, one researcher developed a summary, categorized the article (news, sports, etc), judged if it presented a Black individual or community, if the words “black” or “blacks” were used, and extracted key quotes. This process borrowed the practice of close reading from the fields of literary criticism and text analysis \cite{brummett_techniques_2019}.

Our second approach built on results from the prior analysis to read hand-picked articles that might reveal performance on contemporary issues. We manually selected eight unseen articles in domains we suspected of including false positives and false negatives based on the prior results. All eight articles were analyzed using both quantitative and qualitative assessments as described above.

\section{Results and Discussion}

\subsection{Use of the Label in the NYT Corpus}

The label \textit{blacks} is the 52nd most used label in the training data, appearing on 22,332 articles (1.2\% of the full NYT corpus). The closely related label \textit{blacks (in US)} was the 200th most used label, given to 9,317 articles (0.5\% of the full corpus). 

\begin{figure}
    \centering
    \includegraphics[width=1\linewidth]{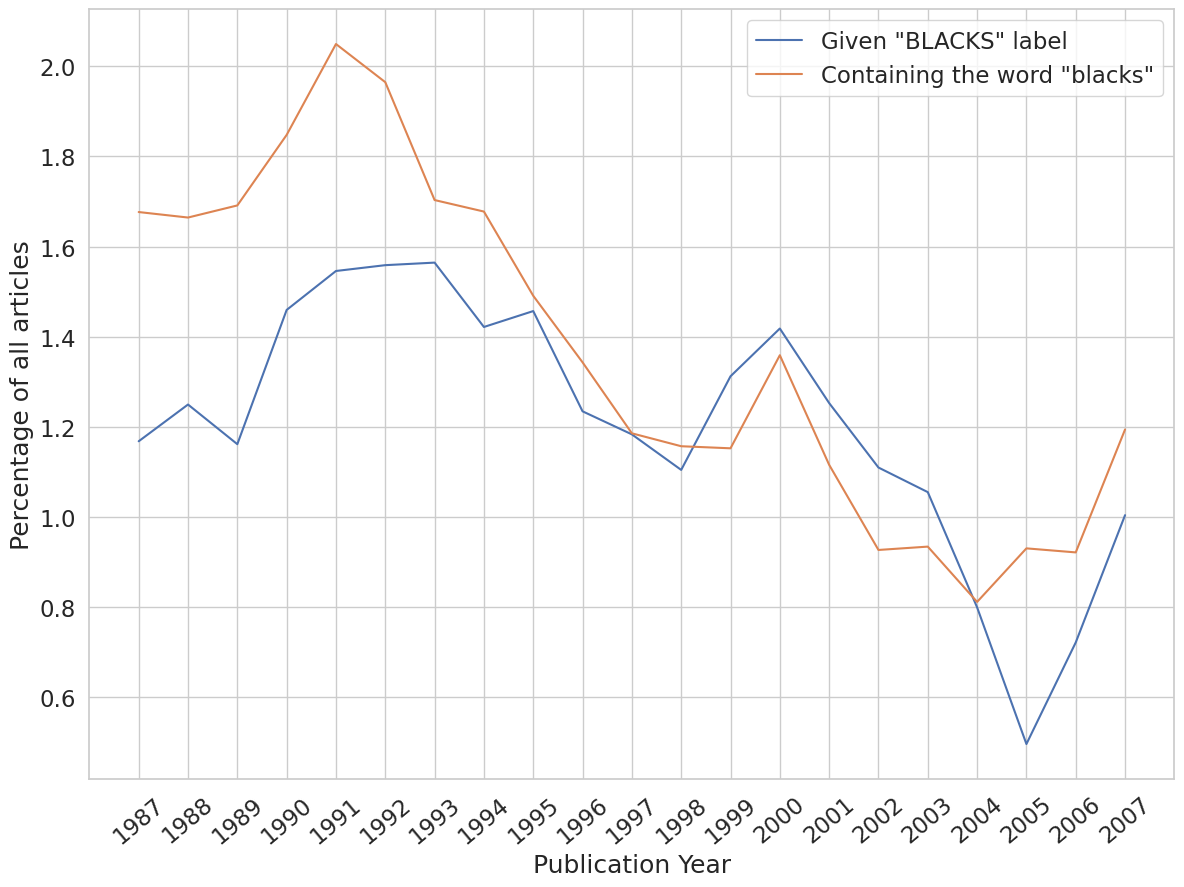}
    \caption{Use of the label \textit{blacks} and the word ”blacks” over time}
    \Description{line chart from 1987 to 2007 showing falling use of blacks label and blacks work over time}
    \label{freq-of-use}
\end{figure}

Figure \ref{freq-of-use} shows the development in the use of the label \textit{blacks} as well as the use of the word "blacks" in news article text. The general decrease in use reflects our simplistic assumptions about social trends in the way Black Americans are referred to in news. Note that the presence of the word "blacks" is highly predictive of the label \textit{blacks}, correlated at 0.82 (p < 0.001). 

We confirmed, by generating a word tree visualizing words in context \cite{wattenberg_word_2008}, that the term (not the label) "blacks" is used in the NYT Annotated Corpus to broadly describe a group of people.

While we focused on the \textit{blacks} label, the most frequently co-occurring labels in the training text can reveal more about the model’s embedded concepts. The top 5 co-occurring labels included: \textit{blacks (in us)}, \textit{politics and government}, \textit{discrimination}, \textit{education and schools}, and \textit{elections}. Many of these reflect and reinforce later findings in the auditing and qualitative review phases.

\subsection{Characterizing the Prediction Space}

\begin{figure}
    \centering
    \includegraphics[width=1\linewidth]{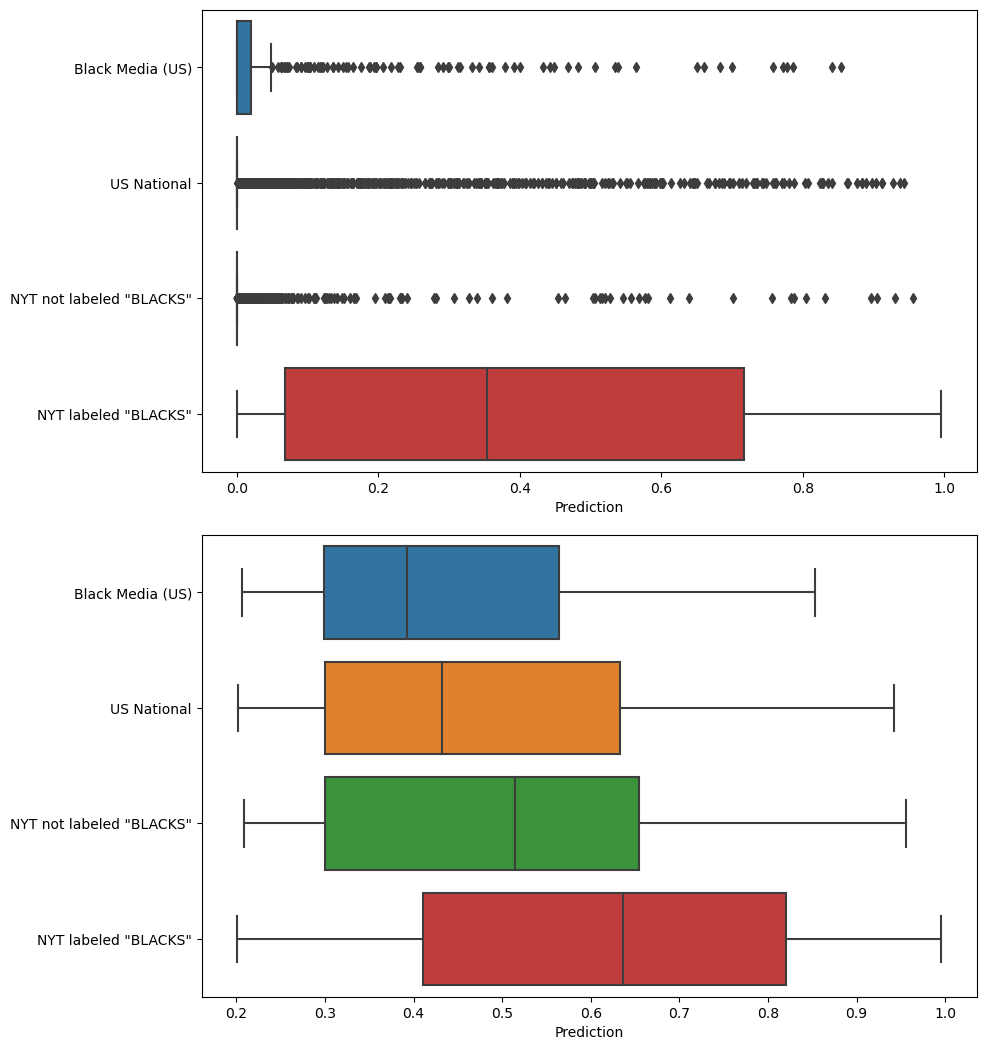}
    \caption{Box-plots showing the distribution of predicted probabilities of being assigned the \textit{blacks} label for each evaluation set. The top plot shows all predictions, the bottom plot shows predictions with probability > 0 for clearer visualization.}
    \Description{box and whisper chart showing distribution of predicted probabilities of being assigned the blacks label for each evaluation set.}
    \label{box-plots}
\end{figure}

Our four evaluation sets offered an opportunity to review application of the label \textit{blacks} in distinct corpora. The distribution of probabilities for each evaluation set can be seen in Figure \ref{box-plots}.

The distributions confirm some hypothesizing, but also raise questions. We anticipated that Set A (NYT labelled \textit{blacks}) would present the highest number of stories scoring above our threshold of 0.2, and that Set C (Black US Media sample) would be second-highest. However, it is surprising that stories labelled as \textit{blacks} (above 0.2) within Set C (Black US Media sample) have lower scores than all the other sets. This suggests some alignment drift between the training corpora (NYT labels and word2vec embeddings) and how authors write in those outlets targeting Black audiences. The small number of stories scoring highly by the model from Set B (NYT not labeled \textit{blacks}) can be best understood as false positives. The distribution of scores in the US National set is best understood as a nominal distribution for overall US media, not revealing much about the model. Overall these charts suggest to us that in usage it is critical to be very intentional about using a thresholded binary label based on a score; the specific value chosen for the threshold can be highly impactful in the performance on unseen content.

\subsubsection{Predictive Terms}

\begin{figure}
    \centering
    \includegraphics[width=1\linewidth]{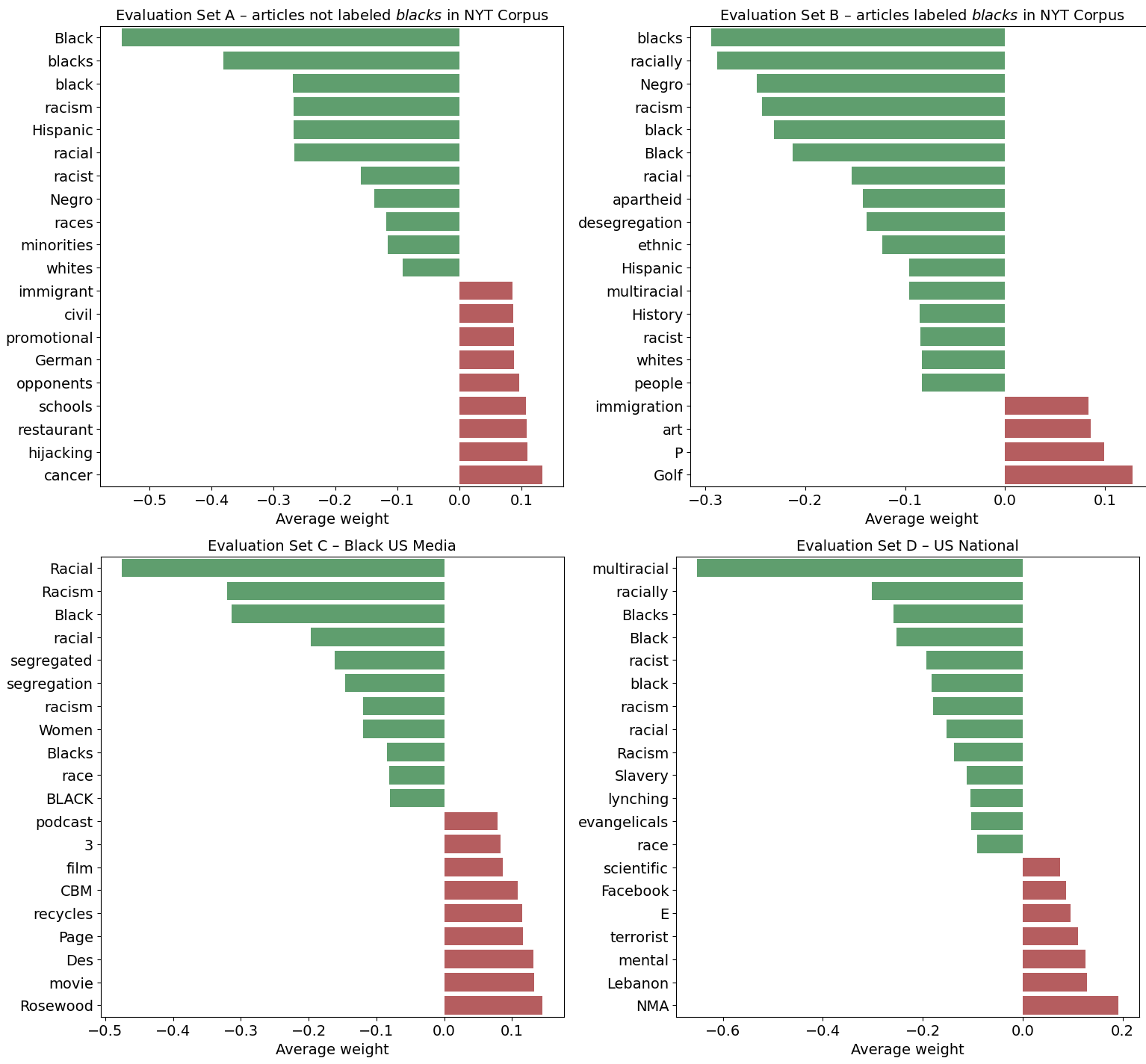}
    \caption{Mean prediction weight of the most influential 10 words across 20 samples from each of the four datasets, as calculated by LIME. Negative weights (green) indicate that the word weighs toward predicting the label \textit{blacks}, positive weights (red) indicate words weighing away from the label.}
    \Description{Composite of four charts, one for each evaluation set, showing the words that are most and least predictive of the blacks label being applied at an above-threshold score.}
    \label{weights}
\end{figure}

In order to understand the potential alignment drift characterized above, we leveraged LIME to determine which terms influenced the prediction of \textit{blacks} in a random sample of 20 articles from each evaluation set. To construct an overall view of trends across articles, for each dataset, this averages the weights of the top ten features of the sampled articles (Figure \ref{weights}).

Words like "racial" and "racism" suggest that articles dealing with topics of racism in one way or another are often labeled \textit{blacks} by the model. This is in accordance with our earlier findings that the label \textit{blacks} co-occurred with the label {discrimination} at 17.34\% and label racial relations at 8.85\% of the time.

We also find that words such as "Hispanic", "minorities", and "women" are also predictive of the label \textit{blacks}. This suggests that news media (via word2vec) and the New York Times (via the annotated corpus) write about these groups in similar ways or in connection with each other. However, Asian Americans, Jewish people, and LGBTQ+ terms do not appear. There can be several reasons: either these groups did not show up in our random samples, they are not written about in the same way as Black Americans, or the concerns and rights of these groups were not on the media agenda in 1987-2007.

\subsection{Content Analysis}

Content analysis of the 25 highest-scoring \textit{blacks}-labeled articles across the four evaluation sets revealed similarities and differences between the training data and contemporary news stories. Here we summarize patterns in the evaluation sets with more focused analysis on a small unseen selection to (a) reveal more about the meaning of the \textit{blacks} label in modern use, and (b) expose gaps in what users might expect it to encode.

\subsubsection{Highest Scored Stories}

The top 25 \textit{blacks}-labeled articles from Set A (training data labeled \textit{blacks} by NYT staff) and Set D (US National) both show a prominent focus on police, law enforcement, juries, and lawsuits. This included stories about diversity among police and specific cases of violence against Black individuals. Racism and discrimination is also a permeating theme. This reflects how \textit{racism/discrimination} and \textit{police/law enforcement} labels co-occurred frequently with \textit{blacks}. Unsurprisingly, the top 25 \textit{blacks}-labeled articles in Evaluation Set B (training data not labeled \textit{blacks} by NYT staff) were more diverse in the topics touched upon. This included coverage of several mentioned topics, with history and education institutions having the highest frequency. The top 25 \textit{blacks}-labeled articles in Evaluations Set C (Black US Media) was likewise more topically diverse. Generally, the model was less certain in its predictions of \textit{blacks} when considering these top 25 articles from Set C as compared to the other datasets.

This close reading of the top 25 \textit{blacks}-labeled articles in each of the four Evaluation Sets reveals additional insights into what the label is encoding. It reinforces prior findings that coverage of Black Americans focuses on negative portrayal such as criminality, victimhood, and poverty \cite{martindale_coverage_1990}. On the other hand, the stories from Evaluation Set C (Black US Media) show fewer policing-related stories, more individual stories, and accounts of systemic harms, perhaps due to editorial norms in that Set that diverge from mass media coverage.

\subsubsection{Asian-American Discrimination}

One early hypothesis was that the \textit{blacks} label was a proxy for stories about racism. This emerged from our analysis of predictive terms, which showed that the \textit{blacks} label was also being applied to stories about other minorities. To understand this more deeply we analyzed two articles about anti-Asian discrimination during COVID-19, one from CNN and one from Fox News. Since the model's training data predates the pandemic, it lacks relevant context for this coverage, and while Asian-related labels existed in the original NYT corpus, they weren't frequent enough to be included in the top-600 model (a fact that merits further study as this omission also raises concerns for the model’s use in journalism settings). The results revealed a "racism detector" problem: the CNN article received a low \textit{blacks} probability (0.04) and was instead labeled as health coverage, while the Fox News article scored 0.35 for \textit{blacks}—above the 0.2 threshold—primarily because it contained the word "racism". A lay person would not likely expect articles about the COVID pandemic to be labeled with \textit{blacks}, yet the Fox News article was. We interpret these findings to suggest the model has encoded use of the word “racism”, no matter the target population, to be something that merits a high \textit{blacks} label score. Stories that discuss the concept of racism are not captured by this label. Thus the \textit{blacks} label is not functioning effectively as a proxy for stories about racism across identity groups.

\subsubsection{Black Lives Matter}

A contemporary example one would expect to receive the \textit{blacks} label would be coverage of the Black Lives Matter (BLM) movement. We similarly selected four stories about BLM, two each from CNN and Fox. Three of the four articles in this sample correctly received the \textit{blacks} label (probabilities ranging from 0.65-0.77), with the word "Black" being the strongest predictor, but one Fox News article about BLM fundraising scored only 0.02 because it primarily used the acronym "BLM" rather than spelling out "Black Lives Matter".

A lay person would expect all four stories about BLM to be labeled \textit{blacks}. The fact that three of the four stories are labeled that way suggests performance matching expectation, but when seeing that the top predictive term is “Black” we become less convinced about the results. Here novel contemporary events don’t seem to be triggering enough of the model’s encoding of historical norms of writing to perform well. The model’s ability to accurately label stories about newer movements relevant to Black people under the label \textit{blacks} is highly sensitive to the language used to describe these movements, and how much it deviates from prior reporting norms. This reveals how the model's historical training data creates gaps related to contemporary terminology (such as abbreviations) that didn't exist when the corpus was created, causing it to miss obviously relevant content simply due to evolving language patterns. Stories about related contemporary Black movements such as \#SayHerName, or the current anti-DEI campaigns of the US government, are examples where this concern about false negatives could occur.

\section{Limitations}

This analysis includes a number of both social and computational limitations that merit discussion. While our research team is diverse across gender and race, none identify as Black American. We believe that there are experiences of being a Black American that would make this work stronger. To mitigate this limitation to some extent we engaged Black researchers and newspaper editors for feedback during the development of this work.

On the technology side, LIME is criticized for including assumptions of linear relationships and feature independence that may not hold for complex transfer learning models \cite{lipton_mythos_2018,hooshyar_problems_2024}. Despite these constraints, we found LIME useful when validated against our domain knowledge and it was just one component of many in our analysis.

A subjective reading led us to believe that articles receiving high \textit{blacks} probabilities appeared predominantly left-leaning in political tone. Our qualitative analysis included both CNN and Fox News articles to partially address this limitation, but broader evidence across diverse partisan media sources would strengthen the validity of these findings.

\section{Conclusion}

The outputs of news-related AI tools risk systemically reproducing historical prejudices that contemporary journalism often seeks to overcome. Our case study investigates the application of the \textit{blacks} label, revealing how out-of-date racial attitudes are embedded in our multi-label classifier trained on the NYT Annotated Corpus. The analysis of specific applications of the label demonstrates the type of feedback loop that can cause decades-old editorial practices to shape present-day applications of AI in story summarization, discovery, audience targeting, and content analysis.

Through correlated labels, predictive terms, and deep reading, we show how the \textit{blacks} label embeds historical conceptions from the training data. Specifically, the label implies that any mention of the word “racism” is somehow connected to Black people, failing to handle social movements that are written about without historical key terms used to refer to those groups. This kind of behavior creates real concerns in the area of how a model creates “representational harms” \cite{gillespie_generative_2024}.

We don’t propose a remedy of abandoning AI tools for work with large news text corpora. Instead we offer that they must be considered as historical artifacts, a statement that requires a hard push against the contemporary marketing mythologies of AI. Practically, we argue that newsrooms and researchers should audit systems before adoption, testing them on contemporary content and examining potential biases in areas that might be sensitive to the specific audiences they serve. Equally important is understanding the training data behind these tools, recognizing that they may not be appropriate for contemporary content analysis. Vendors of commercial products are unlikely to release this information without a broad collective push.

The biases embedded in systems that computational journalists create or build with will shape public understanding of critical issues, making it a moral imperative to ensure these tools support rather than undermine inclusive journalism.

\begin{acks}
This research was supported by the H2020 European Research Council (grant number 834540) as part of the project "The Political Economy of Distraction in Digitized Denmark" (DISTRACT).
\end{acks}

\bibliographystyle{ACM-Reference-Format}
\bibliography{citations}


\end{document}